\documentclass[letterpaper, 10 pt, conference]{ieeeconf}  

\IEEEoverridecommandlockouts                              

\usepackage{amsthm,amsmath,amssymb}
\usepackage{bm}  
\usepackage[ruled,linesnumbered,vlined]{algorithm2e}
\usepackage{mathrsfs}
\usepackage{graphicx}
\usepackage{subfigure}
\usepackage{multirow} 
\usepackage[T1]{fontenc}
\usepackage{balance}
\usepackage{float}
\usepackage{booktabs}  
\usepackage{gensymb}

\title{\LARGE \bf
Indoor Exploration and Simultaneous Trolley Collection  Through Task-Oriented Environment Partitioning
}

\author{Junjie Gao, Peijia Xie, Xuheng Gao, Zhirui Sun, \\Jiankun Wang, \emph{Senior Member, IEEE}, and Max Q.-H. Meng, \emph{Fellow, IEEE}
\thanks{This work is supported by National Natural Science Foundation of China grant \#62103181 and Shenzhen Outstanding Scientific and Technological Innovation Talents Training Project under Grant RCBS20221008093305007. \emph{(Corresponding authors: Jiankun Wang, Max Q.-H. Meng.)}}
\thanks{Junjie Gao, Zhirui Sun, and Jiankun Wang are with the Jiaxing Research Institute, Southern University of Science and Technology, Jiaxing, China.}%
\thanks{All authors are with Shenzhen Key Laboratory of Robotics Perception and Intelligence, and the Department of Electronic and Electrical Engineering of Southern University of Science and Technology in Shenzhen, China.{\tt\small junjie.gao1999@gmail.com}}
}

\begin{document}

\maketitle
\thispagestyle{empty}
\pagestyle{empty}

\begin{abstract}

In this paper, we present a simultaneous exploration and object search framework for the application of autonomous trolley collection. For environment representation, a task-oriented environment partitioning algorithm is presented to extract diverse information for each sub-task. First, LiDAR data is classified as potential objects, walls, and obstacles after outlier removal. Segmented  point clouds are then transformed into a hybrid map with the following functional components: object proposals to avoid missing trolleys during exploration; room layouts for semantic space segmentation; and polygonal obstacles containing geometry information for efficient motion planning. For exploration and simultaneous trolley collection, we propose an efficient exploration-based object search method. First, a traveling salesman problem with precedence constraints (TSP-PC) is formulated by grouping frontiers and object proposals. The next target is selected by prioritizing object search while avoiding excessive robot backtracking. Then, feasible trajectories with adequate obstacle clearance are generated by topological graph search. We validate the proposed framework through simulations and demonstrate the system with real-world autonomous trolley collection tasks.

\end{abstract}

\section{Introduction}
Autonomous mobile manipulation robots have been widely used in domestic and industrial environments to liberate human resources [1], [2]. With the focus on applications of collecting trolleys in indoor environments like airports, our previous work [3], [4] complete the task with prior knowledge of trolley locations. Building on our earlier approaches, we propose a simultaneous exploration and object search framework (shown in Fig. 1), which enables the robot to autonomously explore the unknown space and complete the find-and-fetch task as required in practical situations.

\par Constructed from sensor data, the world representation determines the environment information that can be provided for modules like autonomous exploration and motion planning. However, different information is required to facilitate each system module. Thus, accomplishing a complex task involving multiple modules based on one single map form would be insufficient. To address this issue, we incrementally partition the environment to form a multi-functional world representation. The proposed hybrid map form is composed of labeled room layouts, object proposals, and polygonal obstacles. Consequently, each module can be fed with adequate environment information. 

\par For the exploration module, many approaches have been proposed over the past decades [5], [6], [7]. However, when robots are placed in scenes with complex floor plans, most exploration methods result in revisiting certain rooms due to incomplete previous coverage. This will cause a longer exploration time, especially when robots need to cross multiple rooms through narrow doors. Motivated by this, we extract the LiDAR data belonging to walls to construct room layouts. This can ensure that the robot completes its exploration of the current room before moving to another while prioritizing places of interest using semantic information.

\par For the object detection module, the problem of incomplete camera coverage occurs as a LiDAR sensor is used for efficient environment exploration. Some trolleys may be neglected during the exploration due to the limited field of view of the camera. Therefore, proposals of objects present outside the camera frustum are generated by locating point clusters that are similar in shape and size to a trolley. The remaining points of the LiDAR data are obstacles and represented as polygons. Following our mapping-planning framework in [8], the planner utilizes the geometry and topology information provided by polygons for trajectory generation.

\par The contributions of this paper are summarized as follows:
\begin{itemize}
    \item A novel environment partitioning algorithm, which provides comprehensive information for all system modules. 
   
    \item An exploration-based object search algorithm, which generates the robot's next target by comprehensively considering the spatial distribution of frontiers and potential objects.
    
    \item Simulations and real-world trolley collection tasks are implemented to validate the proposed framework. 
\end{itemize}

\begin{figure*}[t]
\centering
\includegraphics[width=18.0cm,height=6.0139cm]{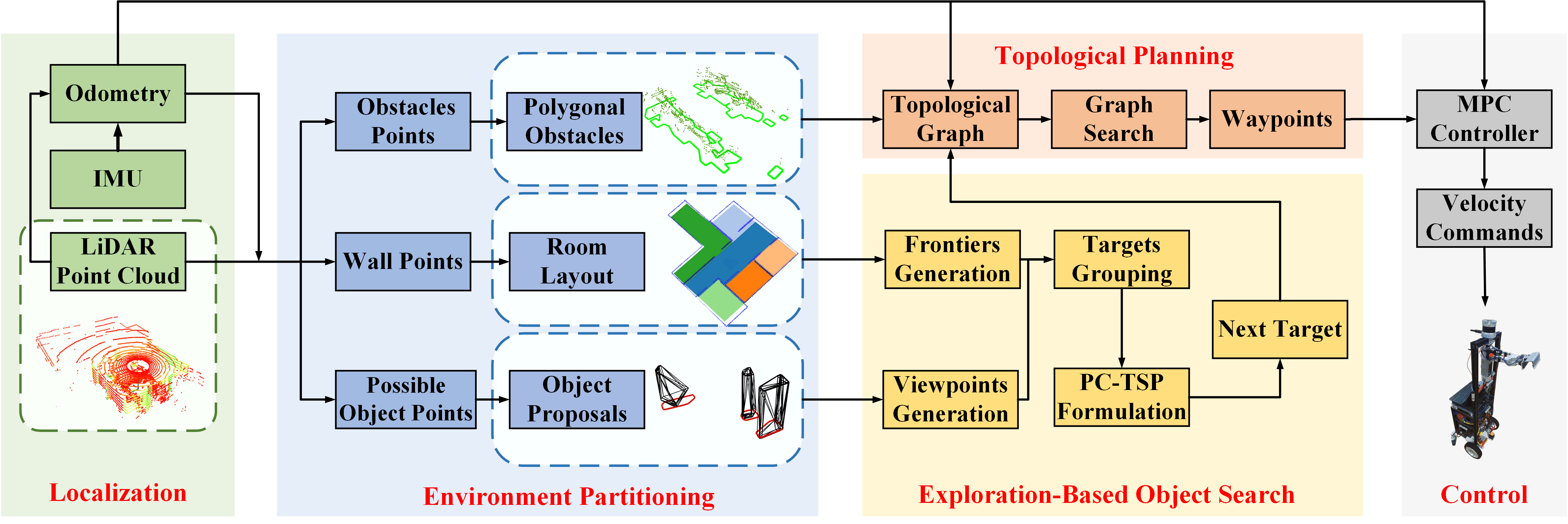}
\caption{Overview of the proposed simultaneous exploration and object search framework.}
\end{figure*}

\section{Related Work}

\subsection{Environment Representation}
For the task of robot motion planning, occupancy grid maps like [9], [10] are commonly used and store the occupancy probability for inquiry. However, the computational cost increases exponentially for more precise descriptions. Moreover, such grids can only provide the information of occupancy status. Thus high-level information of free space and obstacles is necessary to generate feasible trajectories with greater efficiency. As a compact map form. topological maps like [11], [12], [13] describe the connectivity of free space and provide topology information. Although they can greatly speed up the planning task, they still require offline calculations using prior knowledge of the environment. Free space decomposition is also frequently used for convex optimization formulation [14], [15]. In [14], Gao et al. generate convex polygons as flight corridors instead of axis-aligned cubes to capture more free space. To approximate obstacles, polygons and polyhedrons are used in [8], [16], [17] to extract the distribution and geometry information of obstacles to facilitate motion planning. For the task of autonomous exploration, semantic information can significantly reduce the time cost [18], [19]. A semantic road map is built in [18] to incorporate the locations of corridors and rooms. However, such map form is task-specific and the environment information cannot be generalized among different modules of a complex mission.

\subsection{Autonomous Exploration}
Active object search in unknown environments can be accelerated using search heuristics [20], [21]. In [20], Kunze et al. utilize the spatial relationship between objects and related environment structures for fast visual search. However, such spatial relationship is far-fetched in our case, thus the environment needs to be thoroughly explored to collect all trolleys. 
Frontier-based exploration methods is first proposed by Yamauchi et al. in [5]. Defined as boundaries between the known and unknown space, frontiers are extracted as targets using the greedy approach. To improve the exploration efficiency, [7], [22], [23] form a Traveling Salesman Problem (TSP) to select the next target from a large number of frontiers. To facilitate the exploration in complex indoor environments, O{\ss}wald et al. speed up the space coverage using room floor plans in [19]. However, users are required to provide floor plans in advance, and this is impractical for cases in unknown environments. As mentioned earlier, incomplete room coverage during exploration may lead to unnecessary detours. [24] mitigates this problem by applying an incremental map segmentation. However, the applied image segmentation does not necessarily match the real room layout. [25] solves this problem by designing hard constraints for frontier selection using Voronoi graph-based map segmentation, but its greedy approach results in suboptimal traversal order.

\section{Task-Oriented Environment Partitioning}
In this section, we present the proposed environment partitioning method and generate a hybrid world representation $\mathcal{M}$. The environment is divided and transformed into room layouts $\mathcal{M}_{layout}$, polygonal obstacles $\mathcal{M}_{obstacle}$, and potential object proposals $\mathcal{M}_{object}$ to facilitate the execution of the corresponding sub-tasks. The detailed procedures are described as follows:
\subsection{Point Cloud Segmentation} 

We first segment the LiDAR data to extract points belonging to walls and non-architectural parts. 

\par \emph{1) LiDAR Data Preprocessing:} 
As described in Alg. 1, the point cloud $P_{0}$ is first converted to a range image $I_{range}$, and each pixel stores the distance to the measured object. Following [26], the angle matrix $M_{\theta}$ is calculated to describe the tilting angle of the scanned objects. To eliminate points belonging to the floor and ceiling, we perform a bidirectional pixel-wise labeling. Pixels from the lowest and highest rows are labeled as ground clusters $\mathbb{G}=\{G_1,G_2,...,G_m\}$ and ceiling clusters $\mathbb{C}=\{C_1,C_2,...,C_n \}$. Then, their neighbors from different rows are labeled using Breadth-first search (BFS) based on $M_{\theta}$. To cope with the possible sudden changes in the floor and ceiling height, another BFS will start at pixel located at $(r_i, c_i)$ to generate a new cluster in function \textbf{BFSLabel} (line 6, line 8) if the pixel satisfies:
\begin{equation}
\label{eq6}
\left\{
\begin{aligned}
&1 < r_i < I_{range}.rows \\
&\theta_{thresh} < M_{\theta}(r_i, c_i) < \frac{\pi}{2} \\
& |h(r_i, c_i) - h(1, c_i)| < h_{thresh} \\
& |h(r_i, c_i) - h(I_{range}.rows, c_i)| < h_{thresh}
\end{aligned}
\right.
\end{equation}
where $\theta_{thresh}$ is the tilting angle threshold applied in [26], and $h_{thresh}$ is the maximum allowed height change of the floor and ceiling. Next, after eliminating clusters not containing enough points, neighboring clusters with sufficiently small height differences are merged as ground or ceiling. For algorithm output, the rest of the points are denoted as $P_{env}$, and the ceiling height is represented as a piecewise constant function denoted by $h_{ceiling}$.

\begin{algorithm}  
\caption{LiDAR Data Preprocessing}  
\LinesNumbered  
\textbf{Input:} LiDAR data $P_0$ \\
\textbf{Output:} Points for mapping $P_{env}$, height of ceiling $h_{ceiling}$ \\
$I_{range}$, $M_{\theta}$ $\leftarrow$ \textbf{CreateRangeImage}($P_0$)\\
\For{\emph{c = 1}$\dots I_{range}$.cols}
{
    \If {$\neg$  \emph{\textbf{LabeledPixel}(1, c)}}
    {        
        $L$ $\leftarrow$ \textbf{BFSLabel}(1, c, $M_{\theta}$, $\mathbb{G}$)
    }
    \ElseIf{$\neg$  \emph{\textbf{LabeledPixel}($I_{range}.rows$, c)}}
    {
        $L$ $\leftarrow$ \textbf{BFSLabel}($I_{range}.rows$, c, $M_{\theta}$, $\mathbb{C}$)
    }
}
\textbf{RemoveSmallClusters}($L$, $I_{range}$) \\
$h_{ceiling}$, $L$ $\leftarrow$ \textbf{MergeLabels}($I_{range}$, $h_{thresh}$) \\
$P_{env}$ $\leftarrow$ \textbf{ToCloud}($L$, $I_{range}$)\\
\end{algorithm}  

\par \emph{2) Walls and Non-Architectural Components Separation:} 
Then the extracted point set $P_{env}$ is segmented into ${P_{walls}}$ and $P_{non\_arch}$ representing walls and non-architectural components. We make three assumptions to identify walls. First, walls connect the floor and ceiling and are higher than other components. Second, we assume that walls are the boundaries of rooms. Third, we assume that all walls are perpendicular to the floor.

\par Based on assumption 1, a point with height $h_i$ is added to a set ${P_{contour}}$ if it satisfies:
\begin{equation}
h_i \textgreater{\lambda_1  \cdot {h_{ceiling}}}
\end{equation}
where $\lambda_1 \in (0, 1)$ and is set to be lightly less than 1. Then we eliminate points of high non-architectural components like pillars based on assumption 2. The points in ${P_{contour}}$ are divided into different layers by height. As shown in Fig. 2(b), in each layer, a polar coordinate system is generated and the pole $O$ is set as the current LiDAR position. The polar axis is defined using point $A(r_{max},0)$ with the maximum radial coordinate. Starting from $A$, we iterate through all points counterclockwise. Take Fig. 2(b) for example, points with angular coordinates $\alpha \in [\alpha_d, \alpha_e]$ are identified as non-architectural components and removed from ${P_{contour}}$, provided the following equations hold:
\begin{figure}[htbp]
  \centering
  \subfigure[]{
    \label{fig:subfig:onefunction} 
    \includegraphics[scale=0.3]{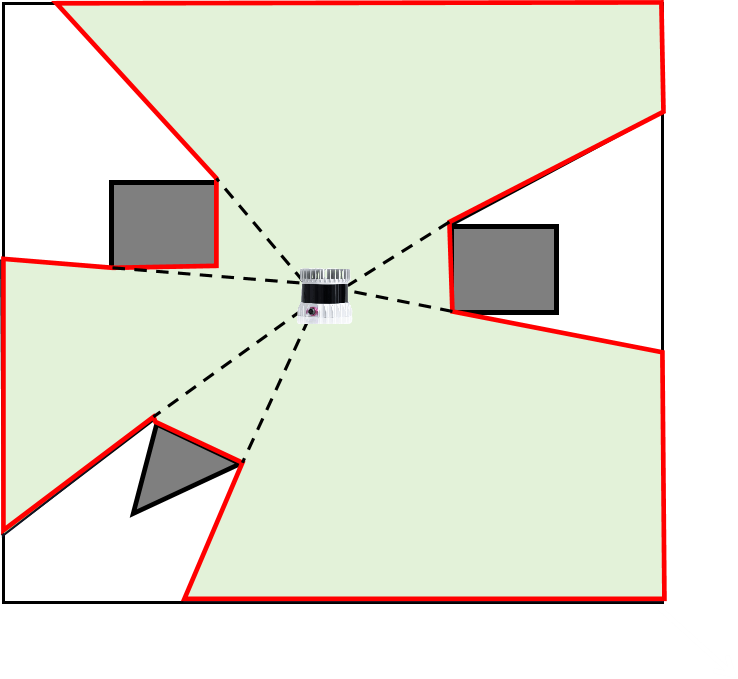}}
  \subfigure[]{
    \label{fig:subfig:onefunction} 
    \includegraphics[scale=0.3]{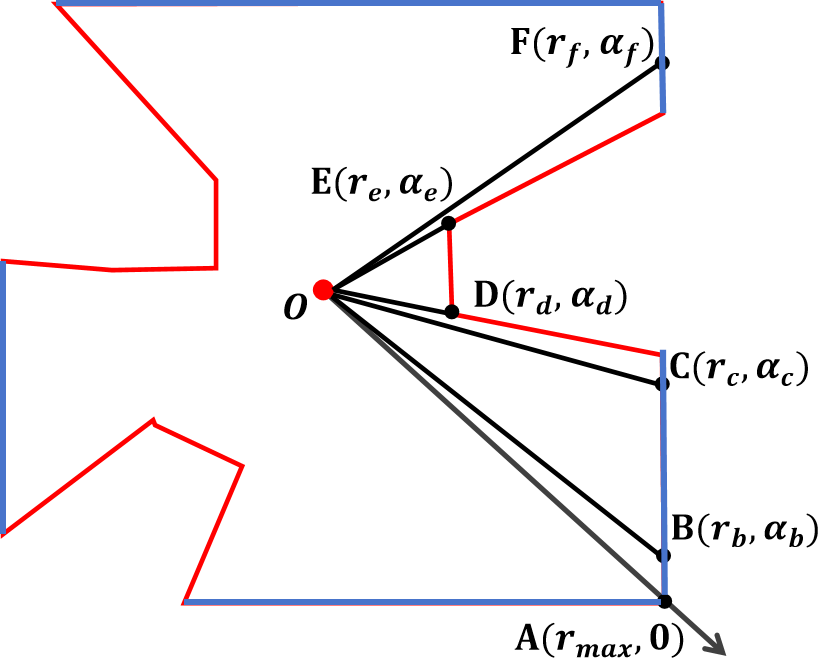}}
  \caption{Illustration of the wall points identification (a) The green polygon and red edges depict the LiDAR coverage area and point cloud distribution, respectively. The walls (the outermost black edges) are not completely covered by LiDAR due to the occlusion of obstacles (gray polygons). (b) A polar coordinate system is built in each layer to divide walls and non-architectural parts. $A, B, C, D, E, F$ are points of the LiDAR data. Point $A$ is used to define the polar axis.}
  \label{fig:twopicture} 
\end{figure}
\begin{equation}
\label{eq6}
\left\{
\begin{aligned}
&r_c - r_d > d_{thresh} \\
&r_e - r_f < -d_{thresh} \\
\end{aligned}
\right.
\end{equation}
where points $C(r_c,\alpha_c), D(r_d,\alpha_d), E(r_e,\alpha_e), F(r_f,\alpha_f)$ are in the same layer and satisfy the condition: $\alpha_c < \alpha_d < \alpha_e < \alpha_f$. Points $C, F$ are neighbors of points $D, E$. $d_{thresh}$ is the threshold to detect abrupt changes in radial coordinates due to obstacle occlusion.

\par With the wall locations derived from the set of partial wall points $P_{contour}$, we then separate the points in $P_{env}$ into $P_{walls}$ and $P_{non\_arch}$ to obtain complete descriptions of walls and non-architectural components based on assumption 3 (shown in Fig. 3). This process takes four steps: In step 1, points in $P_{contour}$ are projected to the floor plane to generate a binary image $I_{binary}$ where white pixels correspond to wall points. In step 2, $I_{binary}$ is blurred and dilated with the robot's size to generate a grayscale image $I_{dilate}$. For step 3, the polygonal wall contour set $\mathbb{W}_{contour}$ is extracted using \emph{OpenCV} [27] function $\bf{findContours}$. In the final step, we iterate through points in $P_{env}$, and a point is added to ${P_{walls}}$ if its projection lies within a polygon $\mathcal{P}_i \in \mathbb{W}_{contour}$. Point set $P_{non\_arch} = P_{env} \verb|\| P_{walls}$. For $\mathcal{P}_i \in \mathbb{W}_{contour}$ we build a KD-tree that narrows down the checking range to points whose projections are inside $\mathcal{O}_i$, where $\mathcal{O}_i$ is the circumcircle of $\mathcal{P}_i$.  

\begin{figure}[htbp]
  \centering
  \subfigure[]{
    \label{fig:subfig:onefunction} 
    \includegraphics[scale=0.13]{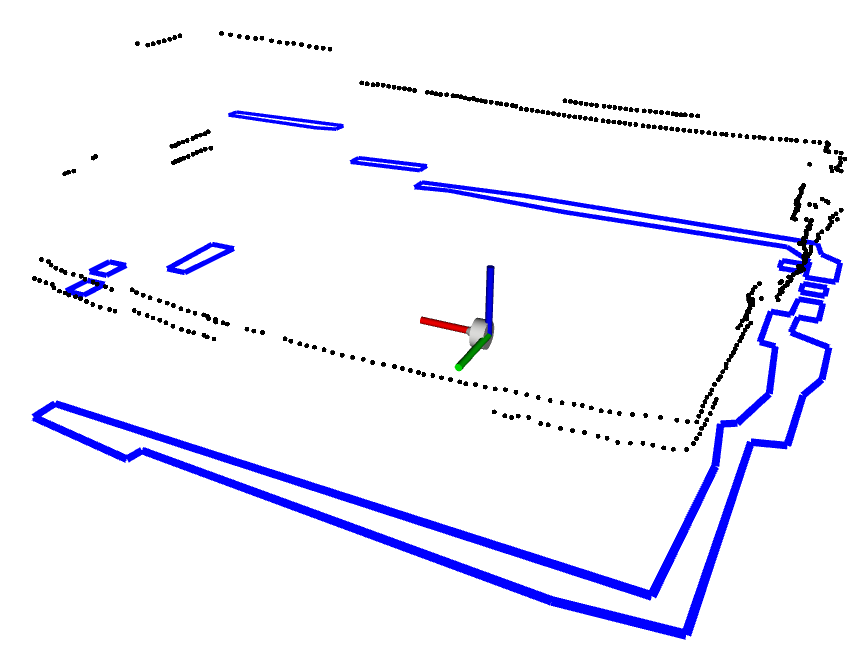}}
  \subfigure[]{
    \label{fig:subfig:onefunction} 
    \includegraphics[scale=0.142]{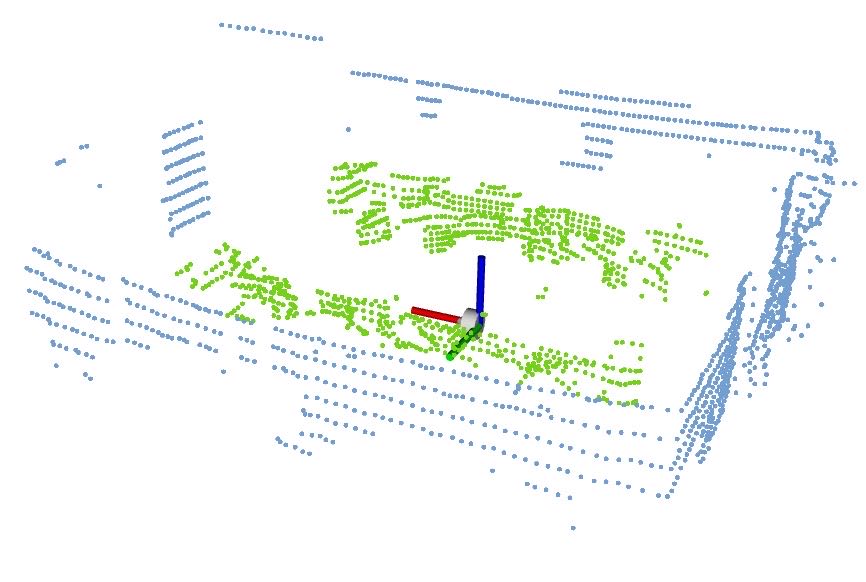}}
  \caption{Demonstration of the point cloud segmentation of one LiDAR data frame (a) Black points are extracted based on our assumptions and represent the upper part of the walls. Blue polygons are their projections. (b) Walls (blue points) and non-architectural components (green points) are thoroughly separated using projections of walls.}
  \label{fig:twopicture} 
\end{figure}

\subsection{Room Layout Generation} 
The room layout $\mathcal{M}_{layout}$ is composed of a line segment set $\mathcal{M}_{layout}^{wall}$ to describe walls and a polygon set $\mathcal{M}_{layout}^{room}$ for room representation. To start with, $P_{walls}$ are used to build a 2D occupancy grid map $\mathcal{M}_o$ [9]. At each iteration, $\mathcal{M}_o$ is converted to a grayscale image $I_{gray}$, which contains three grayscale values corresponding to the occupied grids, unoccupied grids, and unknown grids. Contours of $I_{gray}$ are extracted and denoted as $\mathbb{M}_{contour}$. We compare the vertices of each polygon in $\mathbb{M}_{contour}$ and $\mathbb{W}_{contour}$ to find the modified part of $\mathcal{M}_o$, which is denoted as $\mathcal{M}_{diff}$. Then, we apply the room segmentation methods in [28] and [29] to create the local layout $\mathcal{M}_{layout}^{local}$ from $\mathcal{M}_{diff}$. The detailed process is omitted for brevity. Next,  $\mathcal{M}_{layout}^{local}$ is applied for $\mathcal{M}_{layout}$ update. We find the association between line segments in $\mathcal{M}_{layout}^{wall}$ and  $\mathcal{M}_{layout}^{local}$ by measuring their distance and length to update $\mathcal{M}_{layout}^{wall}$. Newly identified walls are added to $\mathcal{M}_{layout}^{wall}$ as separate line segments. Similarly, polygons in $\mathcal{M}_{layout}^{room}$ are updated by matching their vertices with polygons in $\mathcal{M}_{layout}^{local}$ after deleting polygons that are too small. 

\par In addition, we label each polygon in $\mathcal{M}_{layout}^{room}$ as "separate room" or "corridor" to derive more semantic information. Naturally, corridors are spaces that connect different rooms. Thus polygons that share edges with more than three other polygons are regarded as corridors, while the rest are labeled as "separate room".

\subsection{Object Proposals Generation}
For points in $P_{non\_arch}$, we project them to the floor to obtain the polygonal representation $\mathbb{S}_{non\_arch}$ by extracting contours.
To improve the efficiency of trolley detection and avoid missing objects, we iterate through polygons in $\mathbb{S}_{non\_arch}$ and generate object location proposals $\mathcal{M}_{object}$. First, polygons inside the camera frustum are removed from $\mathbb{S}_{non\_arch}$. Then we traverse $P_{non\_arch}$ to find points whose projections lie within a polygon $\mathcal{P}_j \in \mathbb{S}_{non\_arch}$. Hence for each polygon from $\mathbb{S}_{non\_arch}$, we can obtain a corresponding point cluster $C_j$. $\mathcal{P}_j$ is regarded as a potential object and added to $\mathcal{M}_{object}$ if it satisfies the following conditions:

\begin{equation}
\label{eq6}
\left\{
\begin{aligned}
&s_j \in (s_{thresh}^{lower}, s_{thresh}^{upper}) \\
&v_j \in (v_{thresh}^{lower}, v_{thresh}^{upper}) \\
&I_j^{max} / I_{j}^{mid} \in (I_{thresh1}^{lower}, I_{thresh1}^{upper}) \\
&I_j^{mid} / I_{j}^{min} \in (I_{thresh2}^{lower}, I_{thresh2}^{upper}) 
\end{aligned}
\right.
\end{equation}
where $s_j$ is the area of $\mathcal{P}_j$, and $v_j$ is the volume of the convex hull generated by $C_j$. $I_j^{max}$, $I_j^{mid}$, and $I_j^{min}$ are eigenvalues of the inertia matrix of $C_j$, and $I_j^{max} \geq I_j^{mid} \geq I_j^{min}$ [16]. The above conditions restrict the shape and size of object proposals to be close to that of a trolley. 
Finally, polygons not satisfying Equation (4) are categorized as obstacles and form $\mathcal{M}_{obstacle}$.

\section{Exploration-Based Object Search}
In this section, we present the proposed exploration-based object search algorithm to generate the robot's next target. When no trolley is identified by the visual detection module, a TSP-PC is formulated based on the hybrid map $\mathcal{M}$ to determine the robot's traversal order of all target positions. This allows the robot to implicitly switch the current goal between exploration and trolley collection. The detailed algorithms are described as follows:

\subsection{Potential Targets Generation}
We generate two target sets $T_{exp}$ and $T_{obj}$, which are used to guide the robot to explore unknown places and approach object proposals for inspection, respectively. We also define and store the visiting precedence for different types of targets.

\par \emph{1) $T_{exp}$ Generation:} 
While most frontier-based exploration methods extract frontiers by finding grids adjacent to neighbors with unknown occupancy status, we can directly obtain frontiers from $\mathbb{M}_{contour}$ and $\mathcal{M}_{layout}$. The frontier set $F$ is given by:
\begin{equation}
F =  \max \{\mathcal{P}_k \, | \, \mathcal{P}_k \in \mathbb{M}_{contour}\} - \mathcal{M}_{layout}^{wall}
\end{equation}
where $\mathcal{P}_k$ is a polygon in $\mathbb{M}_{contour}$, and $F$ is a set of line segments. Then we iterate through $F$ and split the line segment in half if its length exceeds a predefined value. $T_{exp}$ is obtained by generating the center point of each line segment in $F$. 

\par \emph{2) $T_{obj}$ Generation:} The robot is described by a bicycle model in our case, and its velocity direction needs to head towards the object proposal for accurate visual detection. Thus for each object proposal $p_i \in \mathcal{M}_{object}$, a viewpoint $v_i$ is generated at a distance $d_{thresh}$ from $c_i$ on the line segment $\overline{c_ir_i}$. $c_i$ and $r_i$ denote the centroid of $p_i$ and the robot's current position, respectively. $d_{thresh}$ is calculated by:
\begin{equation}
d_{thresh} = \lambda_2 \cdot \min\{ d_{range}, d_{\overline{c_ir_i}}\}
\end{equation}
where $\lambda_2 \in (0, 1)$, $d_{range}$ and $d_{\overline{c_ir_i}}$ are the camera sensing range and the distance between $c_i$ and $r_i$, respectively. Pair $(c_i, v_i)$ is added to $T_{obj}$, and $v_i$ needs to be updated using Equation (6) if the robot position changes. $c_i$ is utilized for calculating the traveling cost to other target positions, and $v_i$ is the desired robot position for object proposal inspection. 

\subsection{Precedence-Based Targets Grouping}

After eliminating the targets located inside the polygon in $\mathcal{M}_{obstacle}$, we group the remaining targets based on the following rules:
\begin{itemize}
    \item Elements in ${T}_{obj}$ stand for the potential demands for trolley collection, thus they have the highest priority and are added to a set $T_1$. 
   
    \item If the room where the robot is currently located is labeled as "separate room", then the targets in $T_{exp}$ that lie within this room are added to a set $T_2$.
    
    \item For rooms where the robot is not currently present and labeled as "separate room", the targets in $T_{exp}$ located in these room are added into a set $T_3$.

    \item  Considering that the robot needs to repeatedly pass through corridors when exploring different rooms, targets in rooms labeled "corridor" are given the lowest priority and added to a set $T_4$.
    
\end{itemize}

\subsection{Selection of the Next Target}
We formulate a TSP-PC to calculate the next target based on the aforementioned target grouping. The precedence constraint can be described as: For $i, j \in \{1, 2, 3, 4 \}$ and $i < j$, targets in set $T_i$ must be visited by the robot before the targets in $T_j$. The traveling cost between two targets is defined as the length of a collision-free path, which is generated using the graph search algorithm. Note that the traveling cost from any potential target to the current position is assigned to zero to form a closed-loop tour. Finally, the next target is determined by solving the TSP-PC using the Lin-Kernighan-Helsgaun (LKH) heuristic [30].

\subsection{Topological Motion Planning}
The planning module contains three main steps and is based on our previous work [4], [8]. We first generate a local topological graph [8] in two-dimensional space. During the graph construction, nodes that are in collision with polygonal obstacles are updated by moving in the direction perpendicular to the nearest polygon edge until they stay outside the polygon and maintain enough obstacle clearance. Then we choose the shortest topological path to implement the segmented graph search. In this paper, the Hybrid A* algorithm [32] is used to generate dynamically feasible safe paths. Finally, the executed trajectory is obtained by solving a nonlinear model predictive control (NMPC) problem [4] using waypoints extracted from the previously calculated path.

\par Considering that in some cases the distance between the current robot location and the next target is overly short for topological graph construction, especially when the next target is a viewpoint for object proposal inspection. Therefore, we skip the first step and start directly from the graph search on these occasions.

\section{Experimental Results and Analysis}

\subsection{Mapping Algorithm Validation}
We first analyze the proposed mapping method using our customized mobile robot (shown in Fig. 4(a)). The point cloud is generated by the mounted Ouster OS1 LiDAR sensor with a vertical and horizontal angular coverage of $90^\circ$ and $360^\circ$. We use FAST-LIO [31] to provide odometry information. 
All algorithms are implemented in C++ and integrated through Robot Operating System (ROS). The experiments are conducted by the onboard Intel NUC (specs: Core i7-1165G7 CPU@4.70GHz, 32GB RAM).
\par Mapping parameters are set as Table \uppercase\expandafter{\romannumeral1}, where $I_{thresh1}$, $I_{thresh2}, v_{thresh}, s_{thresh}$ are selected to better approximate the size and shape of a trolley (shown in Fig. 4(b)).

 \begin{figure}[htbp]
  \centering
  \subfigure[]{
    \label{fig:subfig:onefunction} 
    \includegraphics[scale=0.28]{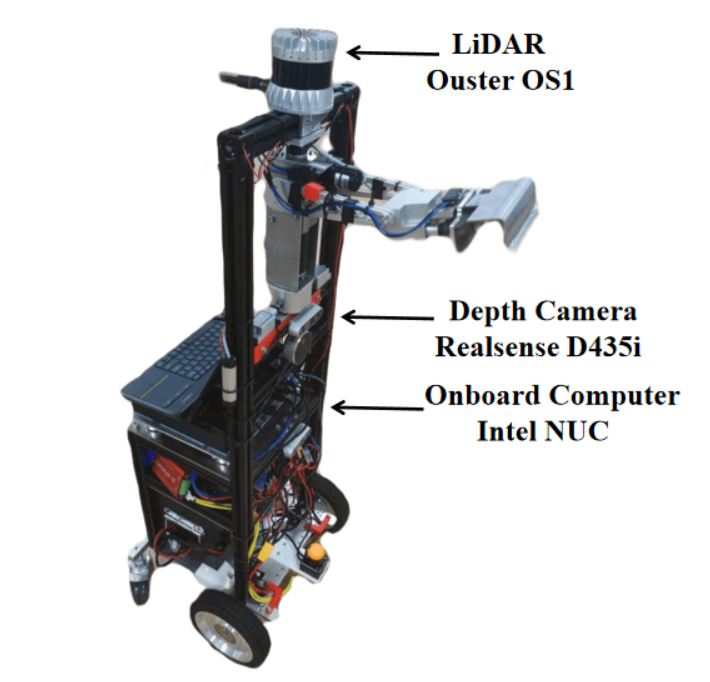}}
  \subfigure[]{
    \label{fig:subfig:onefunction} 
    \includegraphics[scale=0.28]{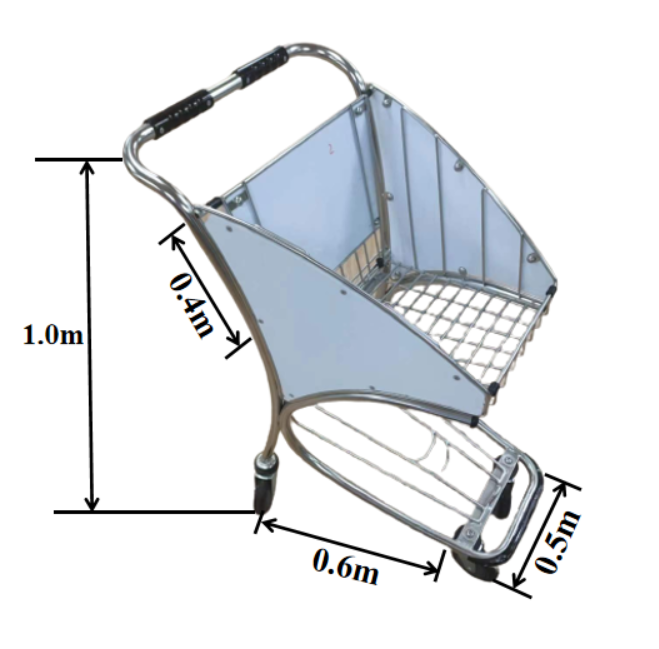}}
  \caption{(a) Hardware setting of our customized robot. (b) The trolley used in our find-and-fetch experiment.}
  \label{fig:twopicture} 
\end{figure}

\begin{table}[H]
\caption{\textbf{Parameters for Environment Representation}}
\centering
\begin{tabular}{|c|c|c|c|c|} 
\hline

Parameter          & Value & Parameter & Value\\ \hline
$\theta_{thresh}$ & $6^\circ$ & $\lambda_1, \lambda_2$ & 0.8 \\ \hline
$d_{thresh}$ & 1.0 $m$ & $s_{thresh}$ &[0.25 $m^2$, 0.5 $m^2$]\\ \hline
$I_{thresh1}$ &[1.0, 1.5]&$I_{thresh2}$ &[1.0, 1.5]\\ \hline
$v_{thresh}$ &[0.25 $m^3$, 0.5 $m^3$]&$h_{thresh}$ &0.5 $m$\\ \hline
\end{tabular}
\end{table}

\par It takes three main steps to transform the LiDAR data into the hybrid map demonstrated in Fig. 1, and the computational time of each step is presented in Table \uppercase\expandafter{\romannumeral2}. We run 20 tests in three environments with different numbers of obstacles and trolleys, and the average time costs for each step are calculated as results. The total time cost proves that the robot can update the hybrid map with a frequency over 18 Hz in indoor environments with dense obstacles, which implies a sufficiently low latency.

\begin{table}[H]
\caption{\textbf{Hybrid Map Construction Run Time}}
\centering
\begin{tabular}{|c|c|c|c|c|} 
\hline

\multirow{3}{2.5cm} {$P_{env}$ Point Number\\/Polygon Number}&Total &{Point Cloud} &{Layout}    &{Object} \\ 
\multirow{3}{*}                                   &Time  &{Segmentation}&{Update}&Identification \\ 
\multirow{3}{*}                                     &(ms)  &(ms)          & (ms)       &(ms) \\ \hline

5384/6 &33.78&8.39&25.12&0.27\\ \hline
7047/10 &45.14&10.82&33.93&0.39\\ \hline
10305/15 &54.19&13.22&40.45&0.52\\ \hline

\end{tabular}
\end{table}

 \begin{figure}[htbp]
  \centering
  \subfigure[]{
    \label{fig:subfig:onefunction} 
    \includegraphics[width=0.45\hsize, height=0.38\hsize]{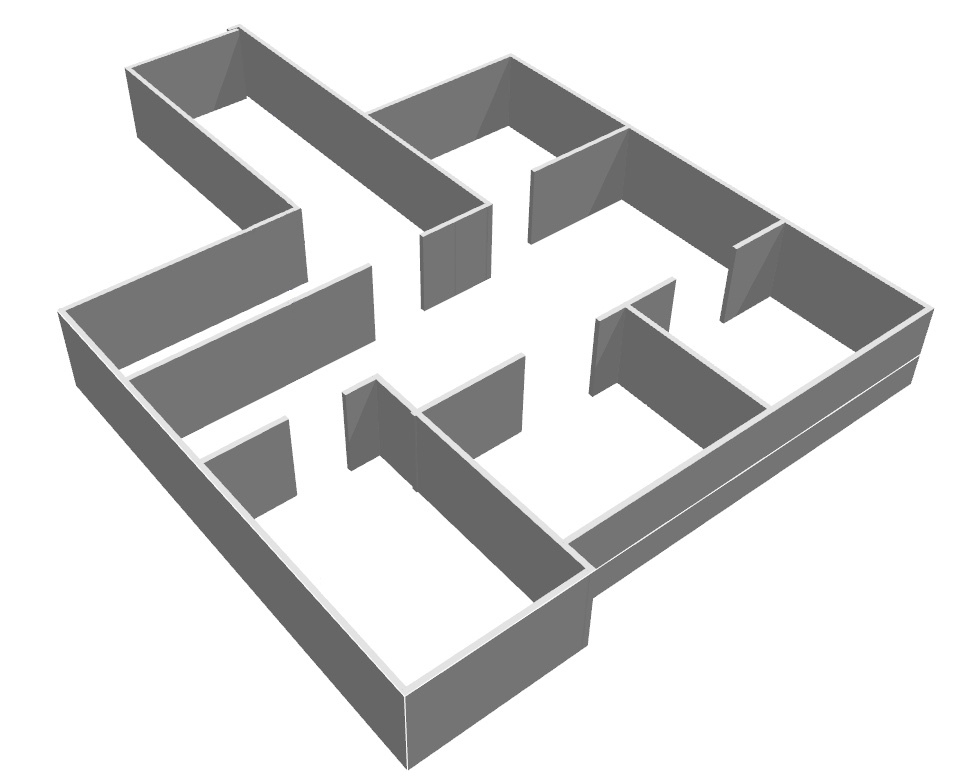}}
  \subfigure[]{
    \label{fig:subfig:onefunction} 
    \includegraphics[width=0.45\hsize, height=0.38\hsize]{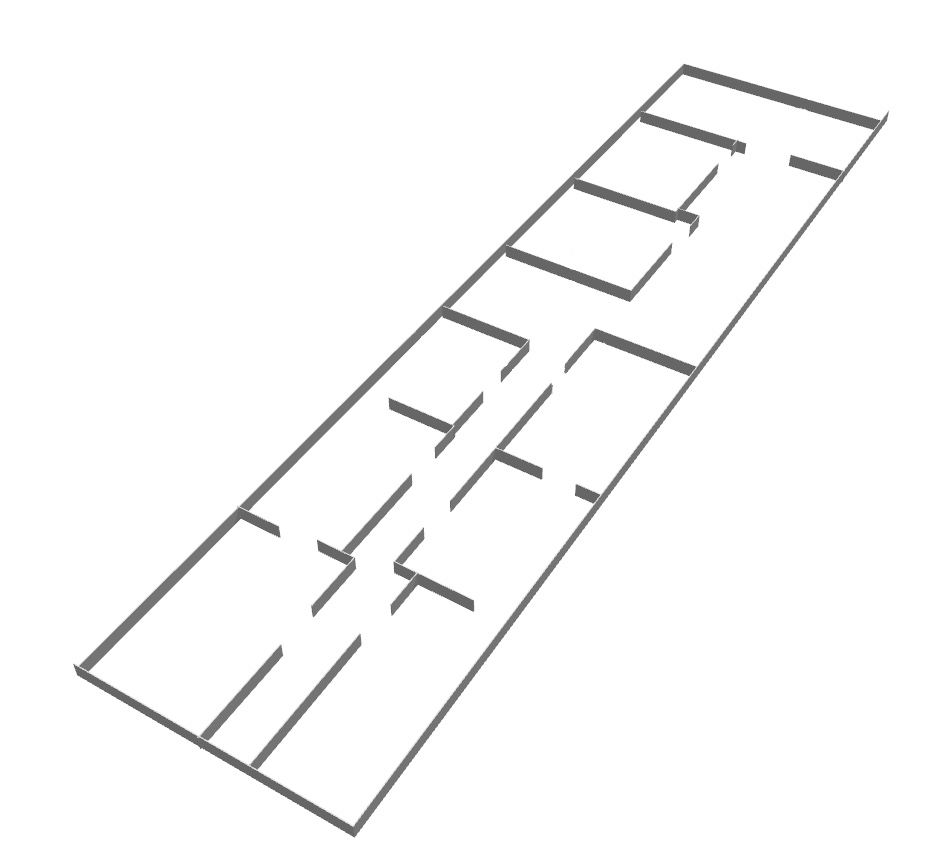}}
  \caption{Gazebo simulation environments (a) The small simulation environment with the size of 20 $m$ $\times$ 20 $m$. (b) The large simulation environment with the size of 40 $m$ $\times$ 100 $m$.}
  \label{fig:twopicture} 
\end{figure}

\subsection{Autonomous Exploration Algorithm Validation}
Then, we validate our autonomous exploration algorithm in simulations. The procedure of object proposal generation is skipped to better compare the performance of the proposed frontier selection method. All simulations are conducted on a laptop with an Intel Core i5-11400H processor. As shown in Fig. 5, we generate two indoor environments with sizes of 20 $m$ $\times$ 20 $m$ and 40 $m$ $\times$ 100 $m$ for robot exploration. The maximum velocity and LiDAR sensing range are set to be 1.5 $m/s$ and 10 $m$, respectively. The generated room layouts with semantic labels of two simulation environments are presented in Fig. 6(a), and the traveled trajectories are shown in Fig. 6(b). We compare our exploration method with the classic frontier-based exploration algorithm [5], FAEL planner [7], and the segmentation-based greedy approach (denoted as seg-based) [25]. Each method is simulated 20 times in both environments, and the average exploration efficiency is demonstrated in Fig. 7. The result shows that our method outperforms other benchmarks in both environments. Although FAEL planner [7] achieves the highest coverage rate at the beginning, detours caused by constantly revisiting the incompletely explored rooms decrease the exploration speed. In addition, the segmentation-based approach [25] reduces the exploration time in both environments compared to the classic frontier-based algorithm [5], which proves the effectiveness of utilizing room layout information. 

 \begin{figure}[htbp]
  \centering
  \subfigure[]{
    \label{fig:subfig:onefunction} 
    \includegraphics[width=0.47\hsize, height=0.42\hsize]{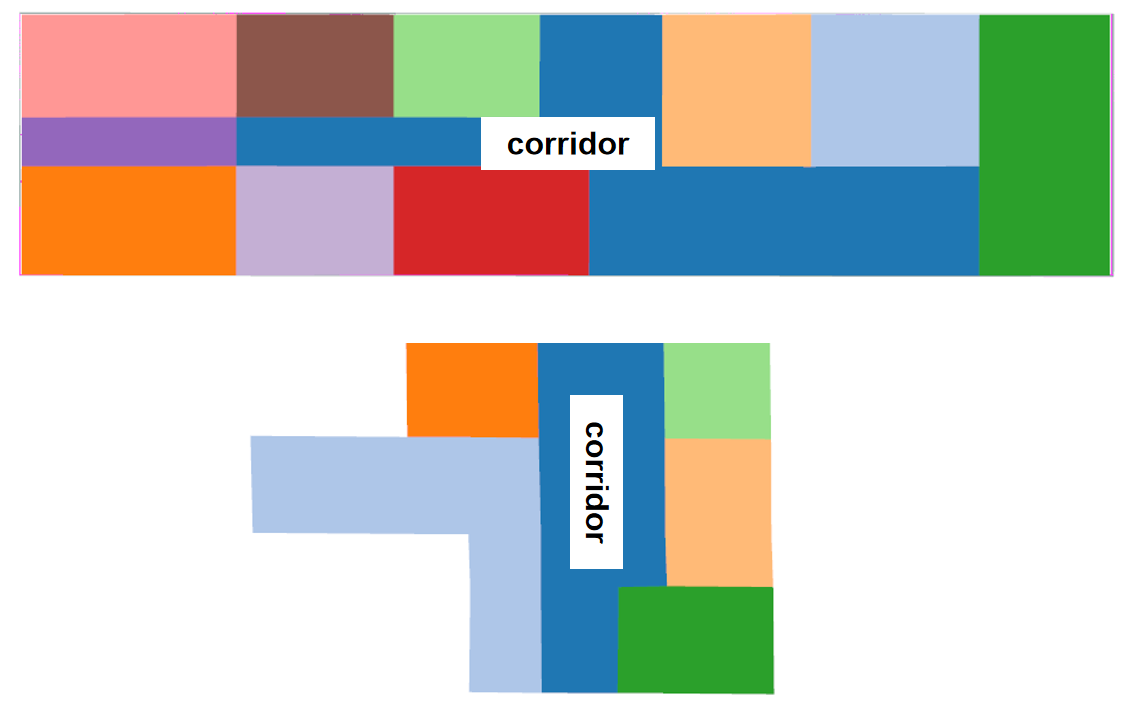}}
  \subfigure[]{
    \label{fig:subfig:onefunction} 
    \includegraphics[width=0.48\hsize, height=0.40\hsize]{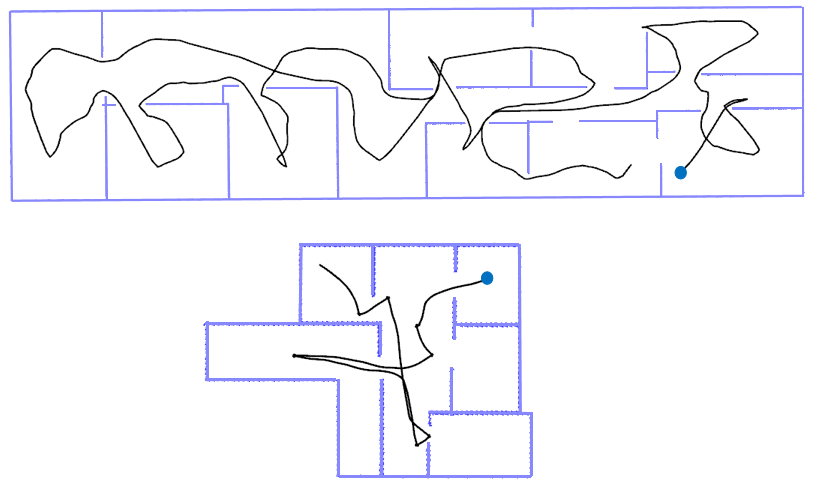}}
  \caption{Autonomous exploration in two simulation environments (a) The constructed layouts with semantic labels. (b) Traveled trajectories in two environments.}
  \label{fig:twopicture} 
\end{figure}

 \begin{figure}[h]
  \centering
  \subfigure[]{
    \label{fig:subfig:onefunction} 
    \includegraphics[width=0.47\hsize, height=0.40\hsize]{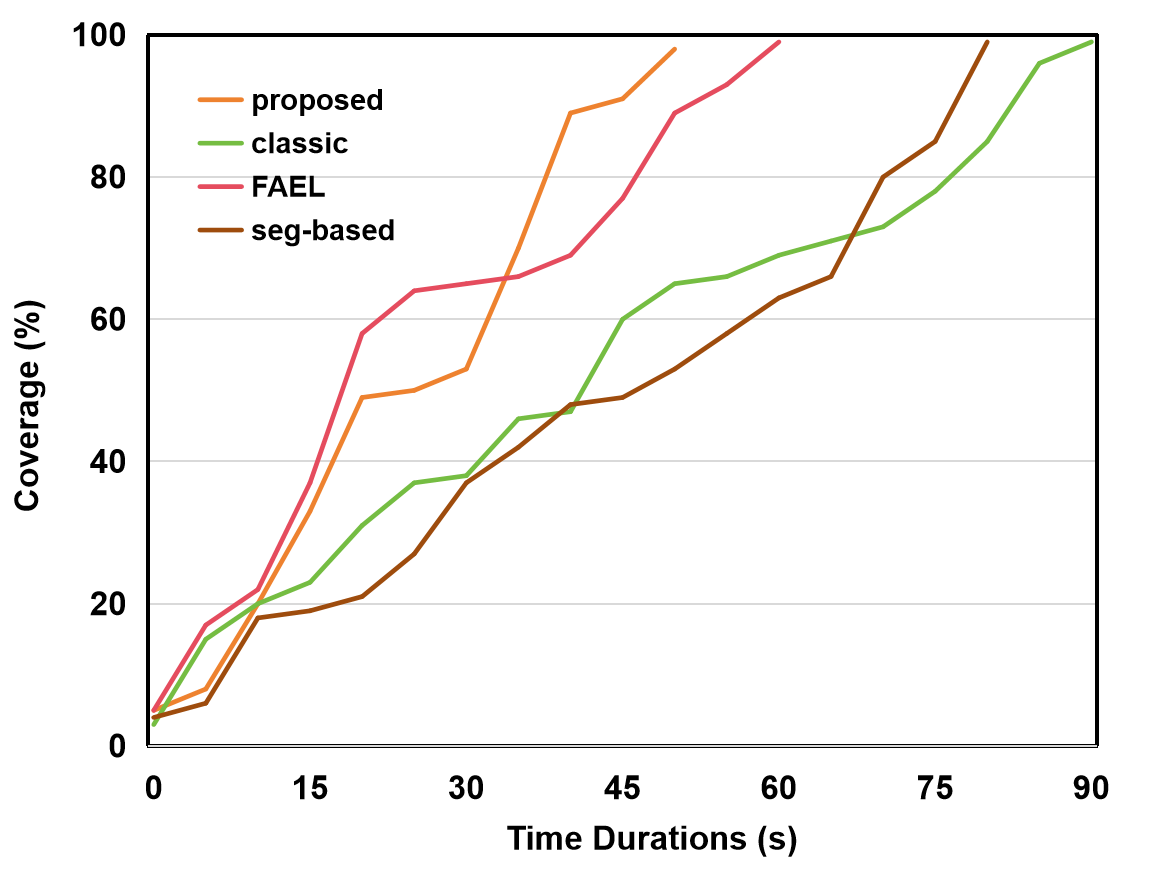}}
  \subfigure[]{
    \label{fig:subfig:onefunction} 
    \includegraphics[width=0.47\hsize, height=0.40\hsize]{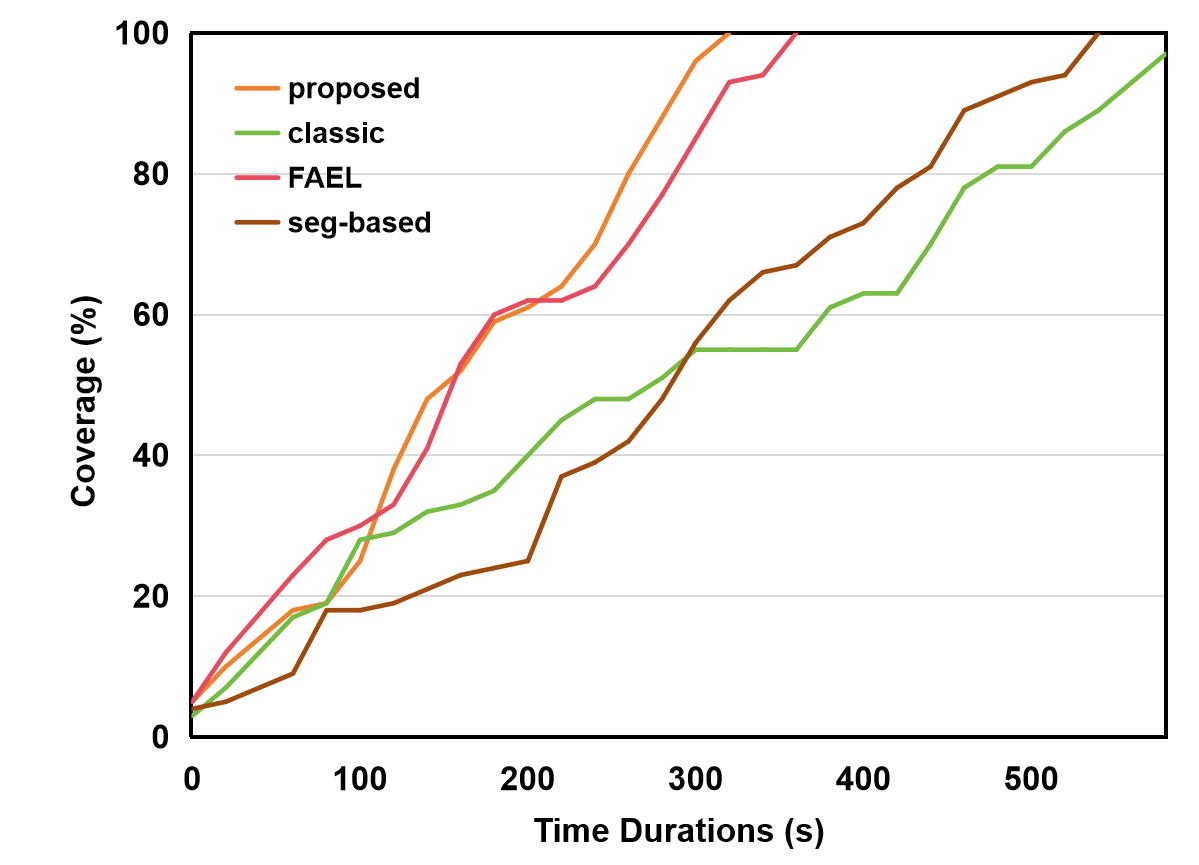}}
  \caption{The comparison of the percentage map coverage against time. (a) The result of simulations in the small environment. (b) The result of simulations in the large environment.}
  \label{fig:twopicture} 
\end{figure}

\subsection{Real-World Trolley Collection Task Demonstration}
Finally, we validate the proposed framework in a real-world autonomous trolley collection task. Starting from the trolley returning spot, the robot is required to fetch all trolleys while exploring the previously unknown indoor environment. 
As shown in Fig. 8, when object proposals are generated during exploration, the robot will proceed toward the selected proposal for inspection. The robot will further approach the object for collection if it is confirmed to be a trolley by the visual detection module. After successfully capturing and returning the trolley, the robot then continues to explore the remaining unknown space. The above process terminates when the exploration is finished. 
\par In our experiments, trolleys are randomly placed during exploration. As shown in Fig. 9, restricted by the robot's current pose, the trolley can not be detected by the camera mounted on the front of the robot. There is no guarantee that the robot will revisit this place later with the desired pose, thus this trolley may not be collected until the task terminates. Such situations verify the necessity of generating object proposals using LiDAR data. 

\begin{figure}[t]
\centering
\includegraphics[width=8.6cm,height=5.05cm]{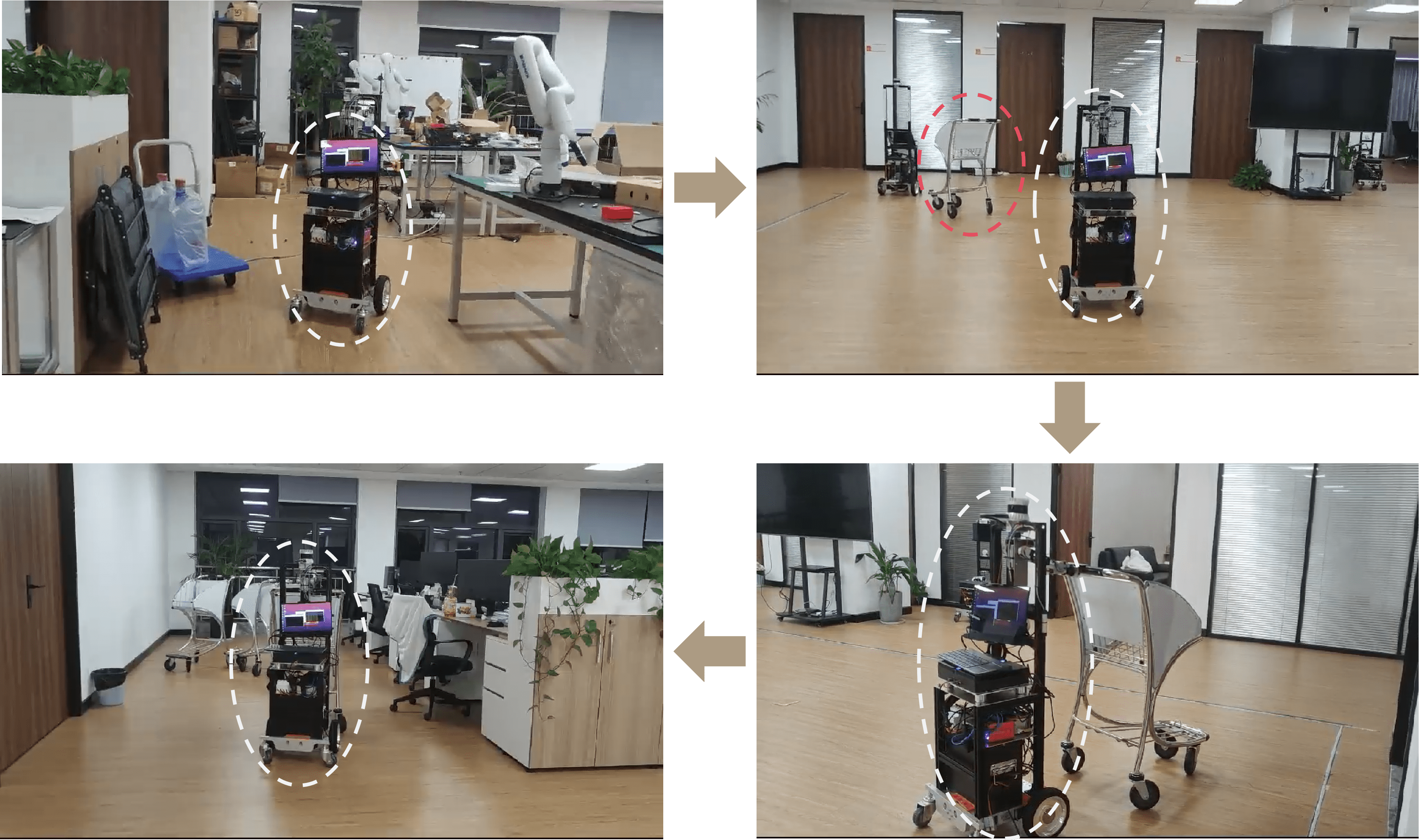}
\caption{Snapshots of the robot (in white circle) conducting the real-world trolley (in red circle) collection task.}
\end{figure}

\begin{figure}[t]
\centering
\includegraphics[width=8.6cm,height=3.05cm]{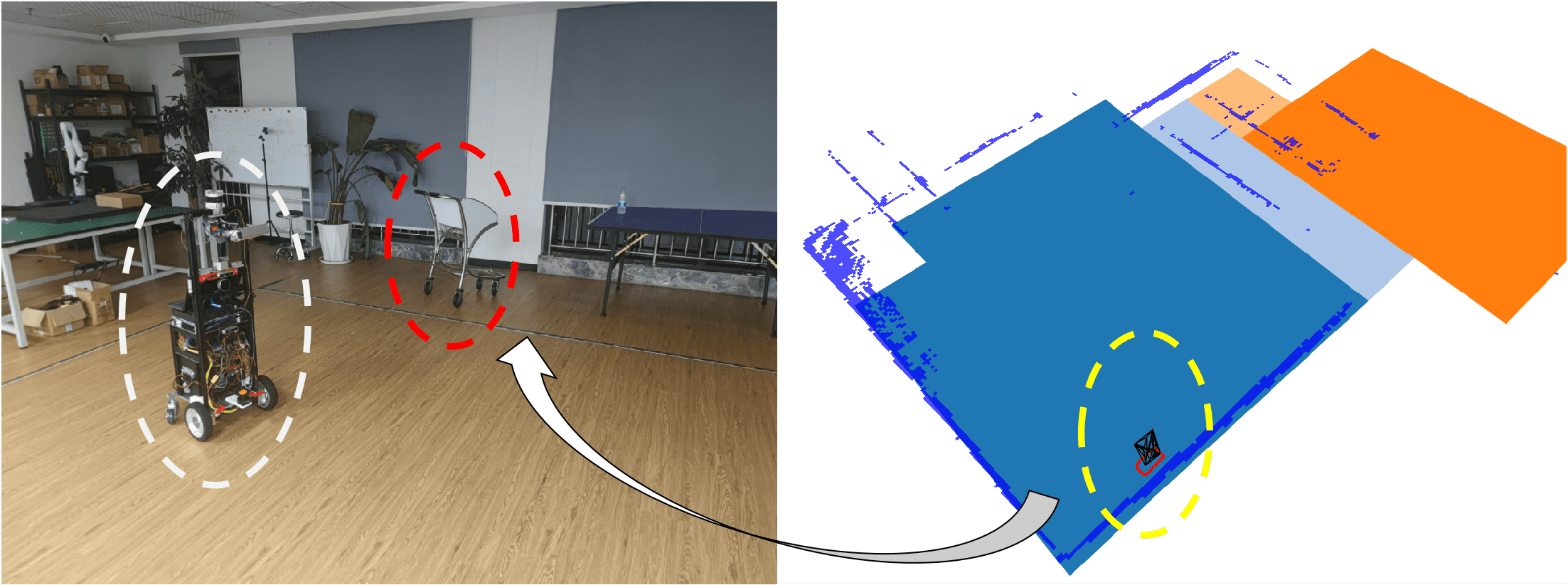}
\caption{The object proposal (in yellow circle) is generated to facilitate the visual detection when the trolley (in red circle) is outside the camera frustum.}
\end{figure}
\section{Conclusions and Future Work}
In this paper, we propose a novel simultaneous exploration and object search framework for autonomous trolley collection in unknown environments. To extract adequate environment information from the sensor data, the LiDAR point cloud is transformed into a hybrid world representation. Composed of room layouts with semantic information, object proposals, and polygonal obstacles, the hybrid map can greatly facilitate the task of autonomous exploration, object search, and motion planning. To generate the robot's next target, the goal of exploration and trolley collection are both considered by forming a TSP-PC, while the object proposals are prioritized for more efficient trolley collection. We demonstrate the proposed framework in both simulation and real-world environments. The experimental results demonstrate that our method outperforms the selected benchmarks and is robust enough for real-world task implementations. Our future work will focus on developing a multi-robot cooperative trolley collection system for applications in large-scale real-world indoor environments.

\addtolength{\textheight}{0cm}   

\balance


\begin{thebibliography}{99}

\bibitem{c1} A. Dömel, S. Kriegel, M. Kaßecker, M. Brucker, T. Bodenmüller, and M. Suppa, “Toward fully autonomous mobile manipulation for industrial environments,” Int. J. Adv. Robot. Syst., vol. 14, no. 4, pp. 1–19, 2017.
\bibitem{c2} Y. Becerra, J. Leon, S. Orjuela, M. Arbulu, F. Matinez, and F. Martinez, “Smart manipulation approach for assistant robot,” in Proc. Int. Conf. Adv. Eng. Theory Appl. Cham Germany: Springer, 2018, pp. 904–913.
\bibitem{c3} C. Wang et al., "Coarse-to-Fine Visual Object Catching Strategy Applied in Autonomous Airport Baggage Trolley Collection," in IEEE Sensors Journal, vol. 21, no. 10, pp. 11844-11857, 15 May15, 2021.
\bibitem{c4} A. Xiao et al., "Robotic Autonomous Trolley Collection with Progressive Perception and Nonlinear Model Predictive Control," 2022 International Conference on Robotics and Automation (ICRA), Philadelphia, PA, USA, 2022, pp. 4480-4486.
\bibitem{c5} B. Yamauchi, “A frontier-based approach for autonomous exploration,” in Proc. IEEE Int. Symp. Comput. Intell. Robot. Automat., 1997, pp. 146–151.
\bibitem{c6} J. M. Pimentel, M. S. Alvim, M. F. Campos, and D. G. Macharet, “Information-driven rapidly-exploring random tree for efficient environment exploration,” J. Intell. Robot. Syst., vol. 91, no. 2, pp. 313–331, 2018.
\bibitem{c7} J. Huang et al., "FAEL: Fast Autonomous Exploration for Large-scale Environments With a Mobile Robot," in IEEE Robotics and Automation Letters, vol. 8, no. 3, pp. 1667-1674, March 2023.
\bibitem{c8} J. Gao, F. He, W. Zhang and Y. Yao, "Obstacle-Aware Topological Planning over Polyhedral Representation for Quadrotors," 2023 IEEE International Conference on Robotics and Automation (ICRA), London, United Kingdom, 2023, pp. 10097-10103.
\bibitem{c9} A. Hornung, K. M. Wurm, M. Bennewitz, C. Stachniss, and W. Burgard, "OctoMap: An efficient probabilistic 3D mapping framework based on octrees," Auto. Robots, vol. 34, no. 3, pp. 189–206, Apr. 2013.
\bibitem{c10} D. Duberg and P. Jensfelt, "UFOMap: An Efficient Probabilistic 3D Mapping Framework That Embraces the Unknown," in IEEE Robotics and Automation Letters, vol. 5, no. 4, pp. 6411-6418, Oct. 2020.
\bibitem{c11} H. Oleynikova, Z. Taylor, R. Siegwart and J. Nieto, "Sparse 3D Topological Graphs for Micro-Aerial Vehicle Planning," 2018 IEEE/RSJ International Conference on Intelligent Robots and Systems (IROS), 2018, pp. 1-9.
\bibitem{c12} F. Blochliger, M. Fehr, M. Dymczyk, T. Schneider and R. Siegwart, "Topomap: Topological Mapping and Navigation Based on Visual SLAM Maps," 2018 IEEE International Conference on Robotics and Automation (ICRA), 2018, pp. 3818-3825.
\bibitem{c13} M. Collins and N. Michael, "Efficient Planning for High-Speed MAV Flight in Unknown Environments Using Online Sparse Topological Graphs," 2020 IEEE International Conference on Robotics and Automation (ICRA), 2020, pp. 11450-11456.
\bibitem{c14} F. Gao, L. Wang, B. Zhou, X. Zhou, J. Pan and S. Shen, "Teach-Repeat-Replan: A Complete and Robust System for Aggressive Flight in Complex Environments," in IEEE Transactions on Robotics, vol. 36, no. 5, pp. 1526-1545, Oct. 2020.
\bibitem{c15} F. Gao et al., "Optimal Trajectory Generation for Quadrotor Teach-and-Repeat," in IEEE Robotics and Automation Letters, vol. 4, no. 2, pp. 1493-1500, April 2019.
\bibitem{c16} D. Wahrmann, A. -C. Hildebrandt, R. Wittmann, F. Sygulla, D. Rixen and T. Buschmann, "Fast object approximation for real-time 3D obstacle avoidance with biped robots," 2016 IEEE International Conference on Advanced Intelligent Mechatronics (AIM), 2016, pp. 38-45.
\bibitem{c17} F. Yang, C. Cao, H. Zhu, J. Oh and J. Zhang, "FAR Planner: Fast, Attemptable Route Planner using Dynamic Visibility Update," 2022 IEEE/RSJ International Conference on Intelligent Robots and Systems (IROS), Kyoto, Japan, 2022, pp. 9-16.
\bibitem{c18} C. Wang, D. Zhu, T. Li, M. Q.-H. Meng and C. W. de Silva, "Efficient Autonomous Robotic Exploration With Semantic Road Map in Indoor Environments," in IEEE Robotics and Automation Letters, vol. 4, no. 3, pp. 2989-2996, July 2019.
\bibitem{c19} S. O{\ss}wald, M. Bennewitz, W. Burgard and C. Stachniss, "Speeding-Up Robot Exploration by Exploiting Background Information," in IEEE Robotics and Automation Letters, vol. 1, no. 2, pp. 716-723, July 2016.
\bibitem{c20} L. Kunze, K. K. Doreswamy, and N. Hawes, “Using qualitative spatial relations for indirect object search,” in Proc. IEEE Int. Conf. Robot. Autom. (ICRA), 2014, pp. 163–168.
\bibitem{c21} C. Wang, J. Cheng, W. Chi, T. Yan and M. Q.-H. Meng, "Semantic-Aware Informative Path Planning for Efficient Object Search Using Mobile Robot," in IEEE Transactions on Systems, Man, and Cybernetics: Systems, vol. 51, no. 8, pp. 5230-5243, Aug. 2021.
\bibitem{c22} C. Cao, H. Zhu, H. Choset, and J. Zhang, “TARE: A hierarchical framework for efficiently exploring complex 3D environments,” Proc. Robot.: Sci. Syst., vol. 5, 2021.
\bibitem{c23} B. Zhou, Y. Zhang, X. Chen, and S. Shen, “FUEL: Fast UAV exploration using incremental frontier structure and hierarchical planning,” IEEE Robot. Automat. Lett., vol. 6, no. 2, pp. 779–786, Apr. 2021.
\bibitem{c24} H. Kim, H. Kim, S. Lee and H. Lee, "Autonomous Exploration in a Cluttered Environment for a Mobile Robot With 2D-Map Segmentation and Object Detection," in IEEE Robotics and Automation Letters, vol. 7, no. 3, pp. 6343-6350, July 2022.
\bibitem{c25} D. Holz, N. Basilico, F. Amigoni and S. Behnke, "Evaluating the Efficiency of Frontier-based Exploration Strategies," ISR 2010 (41st International Symposium on Robotics) and ROBOTIK 2010 (6th German Conference on Robotics), Munich, Germany, 2010, pp. 1-8.
\bibitem{c26} I. Bogoslavskyi and C. Stachniss, “Efficient Online Segmentation for Sparse 3D Laser Scans,” Photogrammetrie - Fernerkundung Geoinformation, vol. 85, pp. 41–52, 12 2016.
\bibitem{c27} Bradski G. The openCV library[J]. Dr. Dobb's Journal: Software Tools for the Professional Programmer, 2000, 25(11): 120-123.
\bibitem{c28} M. Luperto, T. P. Kucner, A. Tassi, M. Magnusson and F. Amigoni, "Robust Structure Identification and Room Segmentation of Cluttered Indoor Environments From Occupancy Grid Maps," in IEEE Robotics and Automation Letters, vol. 7, no. 3, pp. 7974-7981, July 2022.
\bibitem{c29} M. Luperto, V. Arcerito and F. Amigoni, "Predicting the Layout of Partially Observed Rooms from Grid Maps," 2019 International Conference on Robotics and Automation (ICRA), Montreal, QC, Canada, 2019, pp. 6898-6904.
\bibitem{c30} K. Helsgaun, “An effective implementation of the lin–kernighan traveling salesman heuristic,” European journal of operational research, vol. 126, no. 1, pp. 106–130, 2000.
\bibitem{c31} W. Xu and F. Zhang, "FAST-LIO: A Fast, Robust LiDAR-Inertial Odometry Package by Tightly-Coupled Iterated Kalman Filter," in IEEE Robotics and Automation Letters, vol. 6, no. 2, pp. 3317-3324, April 2021.
\bibitem{c32} D. Dolgov, S. Thrun, M. Montemerlo, and J. Diebel, “Path planning for autonomous vehicles in unknown semi-structured environments,” The International Journal of Robotics Research, vol. 29, no. 5, pp. 485–501, 2010.


\end{thebibliography}
\end{document}